\documentclass{ieeeaccess}
\usepackage{cite}
\usepackage{amsmath,amssymb,amsfonts}
\usepackage{algorithmic}
\usepackage{graphicx}
\usepackage{textcomp}
\usepackage{tabularx}
\usepackage{multirow}
\usepackage{makecell}
\usepackage{adjustbox}
\usepackage{url}
\usepackage{empheq}
\usepackage{hyperref}

\def\BibTeX{{\rm B\kern-.05em{\sc i\kern-.025em b}\kern-.08em
    T\kern-.1667em\lower.7ex\hbox{E}\kern-.125emX}}

\begin{document}
	
\history{Date of publication xxxx 00, 0000, date of current version xxxx 00, 0000.}
\doi{10.1109/ACCESS.2021.DOI}

\title{ECG Heartbeat Classification Using Multimodal Fusion}
\author{\uppercase{ZEESHAN AHMAD}\authorrefmark{1}, \IEEEmembership{Graduate Student Member, IEEE},
\uppercase{ANIKA TABASSUM}\authorrefmark{2},\uppercase{LING GUAN}\authorrefmark{3},
\IEEEmembership{Fellow, IEEE}, \uppercase{Naimul Mefraz Khan}\authorrefmark{4}, \IEEEmembership{Senior Member, IEEE}}
\address[1]{ Department of Electrical, Computer and Biomedical Engineering, Ryerson University, Toronto, Canada.(e-mail: z1ahmad@ryerson.ca)}
\address[2]{ Master of Data Science program, Ryerson University, Toronto, Canada. (e-mail: anika.tabassum@ryerson.ca)}
\address[3]{ Department of Electrical, Computer and Biomedical Engineering, Ryerson University, Toronto, Canada.(e-mail: lguan@ee.ryerson.ca)}
\address[4]{ Department of Electrical, Computer and Biomedical Engineering, Ryerson University, Toronto, Canada.(e-mail: n77khan@ryerson.ca)}

\tfootnote{Financial support from NSERC and Dapasoft Inc. (CRDPJ529677-18) to conduct the research is highly appreciated.}

\markboth
{Zeeshan Ahmad \headeretal: ECG Heartbeat Classification Using Multimodal Fusion}
{Zeeshan Ahmad \headeretal: ECG Heartbeat Classification Using Multimodal Fusion}

\corresp{Corresponding author: ZEESHAN AHMAD (e-mail: z1ahmad@ryerson.ca)}

\begin{abstract}
Electrocardiogram (ECG) is an authoritative source to diagnose and counter critical cardiovascular syndromes such as arrhythmia and myocardial infarction (MI). Current machine learning techniques either depend on manually extracted features or large and complex deep learning networks which merely utilize the 1D ECG signal directly. Since intelligent multimodal fusion can perform at the state-of-the-art level with an efficient deep network, therefore, in this paper, we propose two computationally efficient multimodal fusion frameworks for ECG heart beat classification called Multimodal Image Fusion (MIF) and Multimodal Feature Fusion (MFF). At the input of these frameworks, we convert the raw ECG data into three different images using Gramian Angular Field (GAF), Recurrence Plot (RP) and Markov Transition Field (MTF). In MIF, we first perform image fusion by combining three imaging modalities to create a single image modality which serves as input to the Convolutional Neural Network (CNN). In MFF, we extracted features from penultimate layer of CNNs and fused them to get unique and interdependent information necessary for better performance of classifier. These informational features are finally used to train a Support Vector Machine (SVM) classifier for ECG heart-beat classification. We demonstrate the superiority of the proposed fusion models by performing experiments on PhysioNet’s MIT-BIH dataset for five distinct conditions of arrhythmias which are consistent with the AAMI EC57
protocols and on PTB diagnostics dataset for Myocardial Infarction (MI) classification. We achieved classification accuracy of 99.7\% and 99.2\% on arrhythmia and MI classification, respectively. 

Source code at \url{https://github.com/zaamad/ECG-Heartbeat-Classification-Using-Multimodal-Fusion}

\end{abstract}


\titlepgskip=-15pt

\maketitle

\section{Introduction} \label{sec:introduction}
\PARstart{E}{lectrocardiogram} is a reliable, effective and non-invasive diagnostic tool and is the
best representation of electrophysiological pattern of depolarization
and repolarization of the heart muscles during each heartbeat. Heart beat classification based on ECG  provides conclusive information to the cardiologists about chronic cardiovascular diseases~\cite{sun2012ecg}.
An intelligent system for diagnosing cardiovascular diseases is highly desirable because they are the leading source of death around the globe~\cite{xia2017influence}.

Arrhythmia is a heart rhythmic problem which occurs when electrical pulses that coordinate hearbeats cause heart to beat irregularly i.e either too slow or too fast. Arrhythmias can be caused by coronary artery disease, high blood pressure, changes in the heart muscle (cardiomyopathy), valve disorders etc.

\let\thefootnote\relax\footnote{© 2021 IEEE. Personal use of this material is permitted. Permission from IEEE must be obtained for all other uses, in any current or future media, including reprinting/republishing this material for advertising or promotional purposes, creating new collective works, for resale or redistribution to servers or lists, or reuse of any copyrighted component of this work in other works.}

Myocardial Infarction, also known as heart attack, is caused due to the blockage of blood supply to the coronary arteries and in general to the myocardium. This blockage stops the supply of oxygen-rich blood to the heart muscle which can be life-threatening for the patient~\cite{acharya2005study}.

ECG beat-by-beat examination is vital for early diagnosis of cardiovascular conditions. However, differences of recording environment, variations of disease patterns among the subjects during testing, complex, non-stationary and noisy nature of ECG signal~\cite{acharya2018entropies} make heartbeat classification a challenging and laborious exercise for cardiologists~\cite{zhang2014heartbeat}. Thus, computer based novel practices are useful for automatic and autonomous detection of abnormalities in heartbeat ECG classification.

Conventional methods for heartbeat classification using ECG signal rely mostly on hand-crafted or manually extracted features using signal processing techniques such as digital filter-based methods~\cite{pasolli2010active},  mixture of experts methods~\cite{hu1997patient}, threshold-based methods~\cite{chouhan2008threshold}, Principal Component Analysis (PCA)~\cite{bhaskar2015performance}, Fourier Transform~\cite{minami1999real} and wavelet transform ~\cite{khorrami2010comparative}. Some of the classifiers used with these extracted features are Support Vector Machines (SVM)~\cite{sharma2015multiscale}, Hidden Markov Models (HMM)~\cite{chang2012myocardial} and Neural Networks~\cite{lu2000automated}. The first disadvantage with these conventional methods is the separation of feature extraction part and pattern classification part. Furthermore, these methods need expert knowledge about the input data and selected features~\cite{sidek2014ecg}. Moreover, extracting features using subject experts is a time consuming process and features may not invariant to noise, scaling and translations and thus can fail to generalize well on unseen data.

Exemplary performance of deep neural networks (DNNs) on ECG~\cite{krasteva2020fully} and especially the performance of CNN using ID convolution~\cite{tanoh2021novel} and 2D convolution~\cite{wasimuddin2021multiclass} has recently attracted attention of many researchers. Deep learning models are capable of automatically learning invariant and hierarchical features directly from the data and employ end-to-end learning mechanism that takes data
as input and class prediction as output. Recent deep learning models use 1D ECG signal or 2D representation of ECG by transforming ECG signal to images or some matrix form. For 1D ECG classification, commonly used deep learning models are deep
belief networks,  restricted Boltzmann
machines, auto encoders, CNN~\cite{langkvist2014review} and recurrent neural network (RNN)~\cite{salloum2017ecg}. For 2D ECG classification, CNNs are used and the input ECG data is transformed to images or some other 2D representation. It is experimentally proved in~\cite{huang2019ecg} that 2D representation of ECG provides more accurate heartbeat classification compared to 1D. 
In our previous work~\cite{ahmad2020multilevel}, univariate ECG signal is transformed to images by segmenting ECG signal between successive R-R intervals and then stacking these R-R intervals row wise to form images. Finally, multidomain multimodal fusion is performed to improve the stress assessment. Experimental results proved that
multidomain multimodal fusion achieved highest performance  as compared to single ECG modality.

Existing deep learning methods deprived of providing robust fusion framework and rely mostly on concatenation~\cite{dang2019novel} and decision level fusion~\cite{de2004automatic}.

In this manuscript, we deal with the shortcomings of existing deep learning models for ECG heartbeat classification by proposing  two fusion frameworks that have the capacity of extracting and fusing complementary and discriminative features while reducing dimensionality as well.

The proposed work has following significant contributions:

\begin{enumerate}	
	
	\item Two multimodal fusion frameworks for ECG heartbeat classification called Multimodal Image Fusion (MIF) and Multimodal Feature Fusion (MFF), are proposed. At the input of these frameworks, we convert the heartbeats of raw ECG data into three types of two-dimensional (2D) images using Gramian Angular Field (GAF), Recurrence Plot (RP) and Markov Transition Field (MTF). Proposed fusion frameworks are computationally efficient as they keep the size of the combined features similar to the size of individual input modality features.

	\item We transform heartbeats of ECG signal to images using Gramian Angular Field (GAF), Recurrence Plot (RP) and Markov Transition Field (MTF) to conserve the spatial domain correlated information among the data samples. These transformations result in an improvement in classification performance in contrast to the existing approaches of transforming ECG to images using spectrograms or methods involving time-frequency analysis (Short time Fourier transform or wavelet transform).

\end{enumerate}

\section{Related Work}

Deep Learning models especially CNN has been used over the years for ECG heartbeat classification for the detection of cardiovascular diseases such as arrhythmia and MI. These models include both 1D and 2D CNNs.

\subsection{One-dimensional CNN Approaches}

Various models based on 1D CNN has been proposed in the literature for ECG classification. In~\cite{xia2019novel}, an active learning model based on ID CNN is presented for arrhythmia detection using ECG signal. Model performance is improved by using breaking-ties (BT) and modified BT algorithms. Authors in~\cite{kiranyaz2015real} proposed a model for adaptive real time implementation of a patient-specific ECG heartbeat classification based on 1D CNN using end-to-end learning. In~\cite{acharya2017application}, a novel algorithm making use of an 11-layer deep CNN is proposed for automatic detection of MI using ECG beats with and without noise. A transfer learning method based on CNN is proposed in~\cite{kachuee2018ecg} where the information learned from arrhythmia classification task is employed as a reference for the training of classifiers. A computationally intelligent method for patient screening and arrhythmia detection using CNN is proposed in~\cite{pourbabaee2018deep}. The proposed method is capable of diagnosing arrhythmia conditions without expert domain knowledge and feature selection mechanism. In~\cite{tripathy2019localization}, wavelet transform based on Fourier-Bessel series expansion
is proposed for the localization of ECG. The Fourier-Bessel spectrum of the ECG beats is separated into adjacent parts using the fixed order ranges and then multiscale CNN is employed for MI classification of different categories. 
Multi-Channel Lightweight Convolutional Neural Network (MCL-CNN) which uses squeeze convolution, the depth-wise convolution, and the point-wise convolution is proposed in~\cite{chen2018multi} for MI classification.
Two end-to-end deep learning models based on CNN are proposed in~\cite{shaker2020generalization}. These models are called two stage hierarchical model. Furthermore, generative adversarial networks (GANs) is used for data augmentation and to reduce the class imbalance.
In~\cite{wang2020high}, authors proposed a neural network model for precise classification of heartbeats by following the AAMI inter-patient standards. This model works in two steps. In the first step the signals are preprocessed and then features are extracted from the signals. In the second step, the classification is performed by a two-layer classifier in which each layer consists of two independent fully-connected neural networks. The experiments show that the proposed model precisely detects arrhythmia conditions. 
In~\cite{chen2020automated}, authors proposed a complex deep learning model consists of CNN and LSTM. This model classifies six types of ECG signals by processing ten seconds ECG slices of MIT-BIH arrhythmia dataset. Experimental results proved that the proposed model could be used by cardiologists to detect arrhythmia. 
In~\cite{porumb2020convolutional}, authors presented CNN based model for proper diagnoses of  congestive heart failure using ECG. The testing and training of the proposed model was carried out on publicly available ECG datasets. Performance of the proposed model shows the authenticity of model for congestive heart failure detection.

\subsection{Two-dimensional CNN Approaches}

The knock out performance of CNN on 2D data such as images convinced the researchers to convert raw ECG data to images for improved results. In~\cite{huang2019ecg}, short-time Fourier transform is used to convert ECG signal into time-frequency spectrograms that were used as input to CNN for arrhythmia classification. Experimental results show that 2D-CNN achieved higher classification accuracy than 1D-CNN. In~\cite{hao2019spectro}, ECG signal is converted  into spectro-temporal images that were sent as an input to multiple dense convolutional neural network to capture both beat-to-beat and single-beat information for analysis. Authors in~\cite{oliveira2019novel} transformed heartbeat time intervals of ECG signals to images using wavelet transform. These images are used to train a six layer CNN for heartbeat classification. In~\cite{al2019dense}, Generative neural network is used to convert the raw 1D ECG signal data into a 2D image. These images are input to DenseNet which produces highly accurate classification, with high sensitivity and specificity using 4 classes of heart beat detection. To distinguish abnormal ECG samples from
normal, authors in~\cite{diker2019novel} used pretrained CNNs such as AlexNet, VGG-16 and
ResNet-18 on spectrograms obtained from ECG. Using a transfer learning approach, the highest accuracy of 83.82\% is achieved by AlexNet. In~\cite{liu2017real}, multi-lead ECG are treated as 2D matrices for input to a novel model called multilead-CNN (ML-CNN) which employs sub two-dimensional (2D) convolutional layers and lead asymmetric pooling (LAP) layers. 
In~\cite{zhai2018automated}, authors generated dual beat coupling matrix from the sections of heartbeats.  This dual beat coupling matrix was then as 2D input to a CNN classifier. Gray-level co-occurrence matrix (GLCM), obtained from ECG data is employed for features vector description due to its exceptional statistical feature extraction ability in~\cite{sun2019morphological}. In~\cite{izci2019cardiac}, ECG signals were segmented into heartbeats and each of the heartbeats were
transformed to 2D grayscale images which were input to CNN. 
In~\cite{mathunjwa2021ecg}, two second segments of ECG signal are transformed to recurrence plot images
to classify arrhythmia in two steps using deep learning model. In the first step the noise and ventricular fibrillation (VF) categories were recognized and in the second step, the atrial fibrillation (AF), normal, premature AF, and premature VF labels were classified. Experimental results show the promising performance of the proposed method.

\subsection{Fusion based approaches}

Fusing different modalities mitigates the weaknesses of individual modalities both in 1D and 2D forms by integrating complementary information from the modalities to perform the analysis and classification
tasks accurately. In~\cite{fan2018multiscaled}, a Multi-scale Fusion convolutional neural network (MS-CNN) is proposed for heartbeat classification using ECG signal. The Multi-scale Fusion convolutional neural network is a two stream network consisting of 13 layers. The features obtained from the last convolutional layer are concatenated before classification. Another Deep Multi-scale Fusion CNN (DMSFNet) is proposed in~\cite{wang2020deep} for arrhythmia detection. Proposed model consists of backbone network and two different scale-specific networks. Features obtained from two scale specific networks are fused using a spatial attention module. Patient-specific heartbeat classification network based on a customized CNN is proposed in~\cite{li2019automated}. CNN contains an important module called  multi-receptive field spatial feature extraction (MRF-SFE). The MRF-SFE module is designed for extracting multispatial deep features of the heartbeats using five parallel convolution layers with different receptive fields. These features are concatenated before being sent to the third convolutional layer for further processing. Two stage serial fusion classifier system based on SVM’s rejection
option is proposed in~\cite{uyar2007arrhythmia}. SVM’s distance outputs are related with confidence measure and then ambiguous samples are rejected with first level SVM classifier. The rejected samples are then forwarded to a second stage Logistic Regression classifier and then late fusion is performed for arrhythmia classification. 
Authors in~\cite{zhao2017electrocardiograph} presented a unique feature fusion method called parallel graphical feature fusion where all the focus is given to geometric features of data. Original signal was
first split into subspaces, then multidimensional features are extracted from these subspaces and then
mapped to the points in high-dimensional space.
Multi-stage feature fusion framework based on CNN and attention module was proposed in~\cite{wang2019multi} for multiclass arrhythmia detection. Classification is performed by extracting features from different layers of CNN. Combination of CNN and the attention module shows the improved discrimination power of the
proposed model for ECG classification.

The shortcoming in the existing fusion methods is that they depend mostly on concatenation fusion. Concatenation leads towards the problem computational complexity, curse of dimensionality and hence the degradation in classification accuracy~\cite{manshor2011feature}.
In this paper, we address the imperfections of the existing literature and propose two fusion frameworks called Multimodal Image Fusion (MIF) and Multimodal Feature Fusion (MFF) which extract and fuse the features while reducing dimensionality as well. The proposed fusion frameworks are described in section~\ref{sec:proposed method}.
\begin{figure*}
	\centering
	\includegraphics[width=0.9\linewidth]{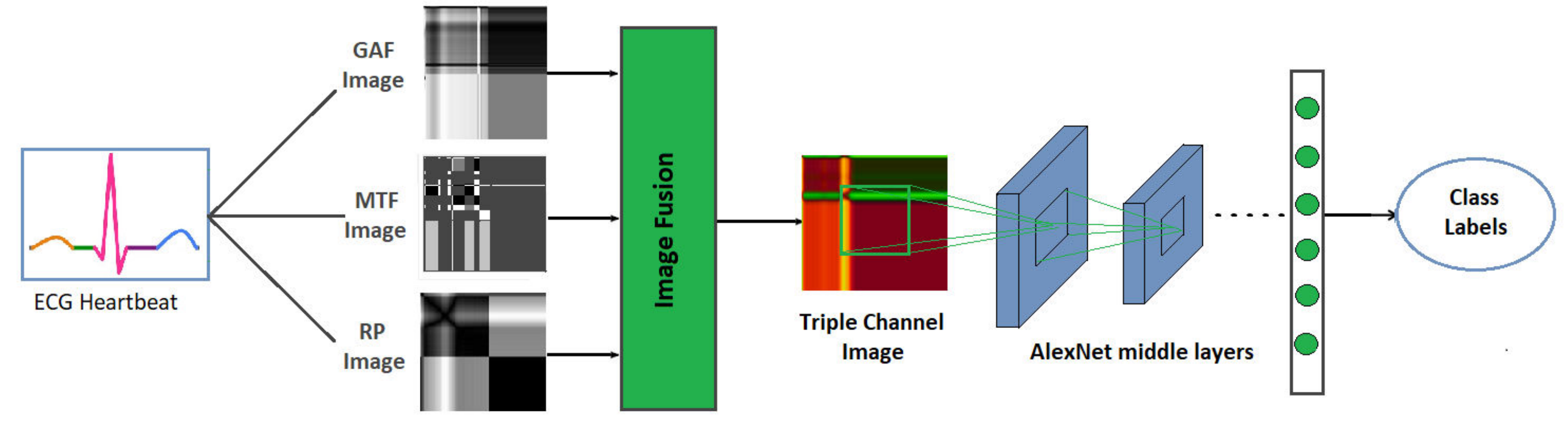}
	\caption{Complete Overview of the Proposed Multimodal Image Fusion (MIF) Framework. We fused GAF, RP and MTF image to form a triple-channel (GAF-RP-MTF) compound image containing both static and dynamic features of input images.}
	\label{fig: MIF framework}
\end{figure*}
\begin{figure*}
	\centering
	\includegraphics[width=0.9\linewidth]{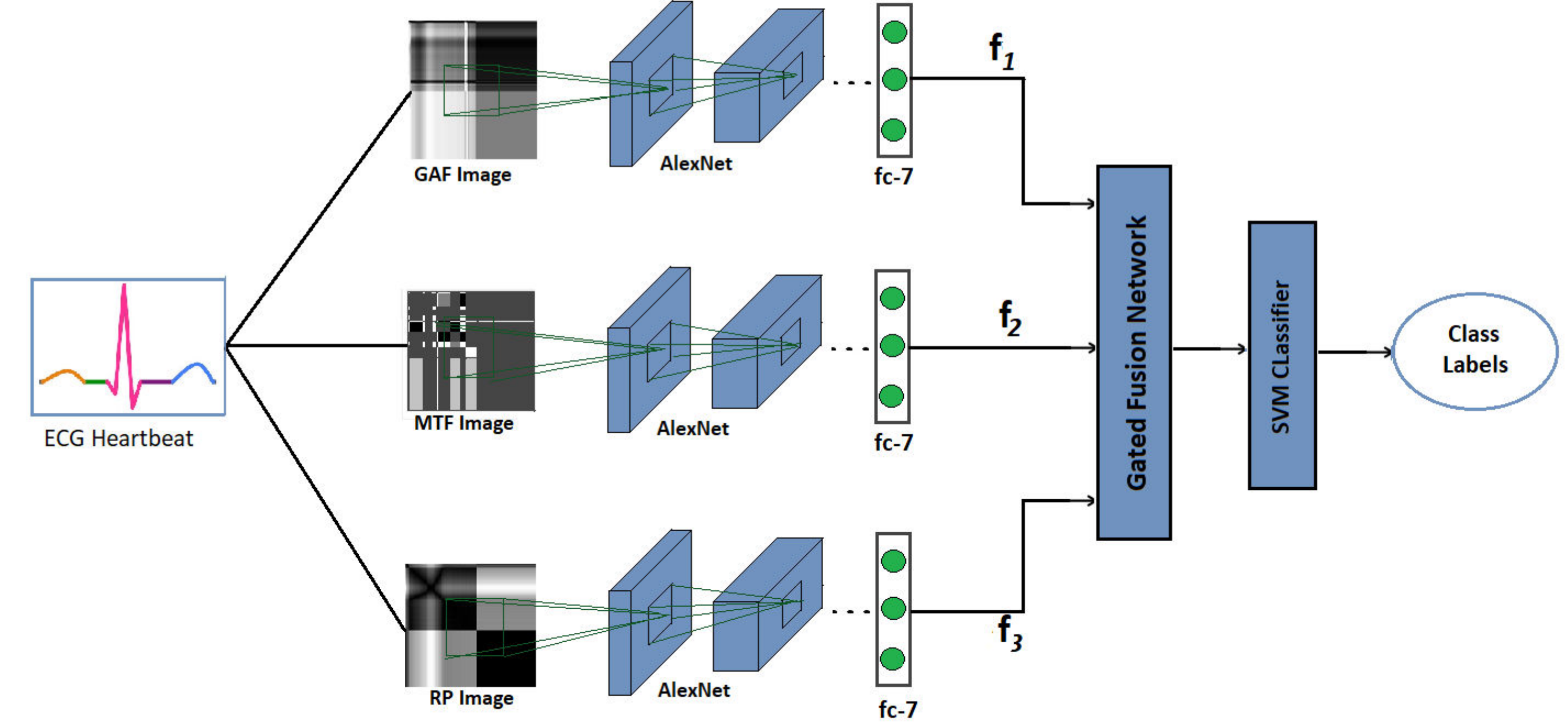}
	\caption{Complete Overview of the Proposed Multimodal Feature Fusion (MFF) Framework. The MFF extracted features
		from fc-7 layer of AlexNet. These features are then integrated through Gated Fusion Network (GFN)
		and are finally sent to the classifier}
	\label{fig: MFF framework}
\end{figure*}

\begin{figure*}
	\centering
	\includegraphics[width=0.8\linewidth]{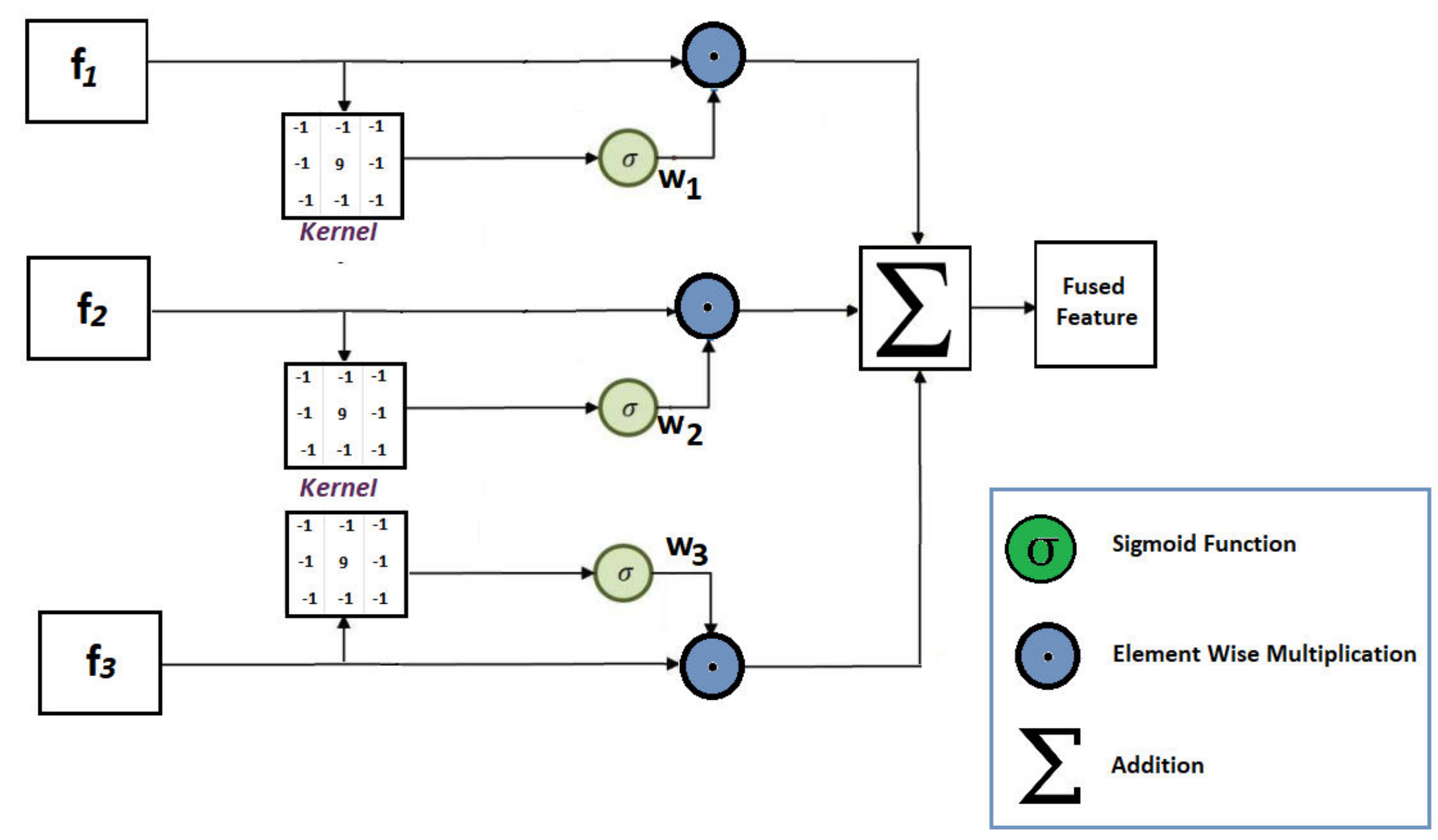}
	\caption{Structure of the proposed Gated Fusion Network. Input feature $f_1$, $f_2$ and $f_3$ from modalities are convolved with high boost kernel and then gated values  $w_1$, $w_2$ and $w_3$ are generated using sigmoid function. Finally, these gated values are multiplied element-wise with input features to perform fusion.}
	\label{fig: gated fusion network}
\end{figure*}
\begin{figure*}[h]
	\centering
	\includegraphics[width=\linewidth]{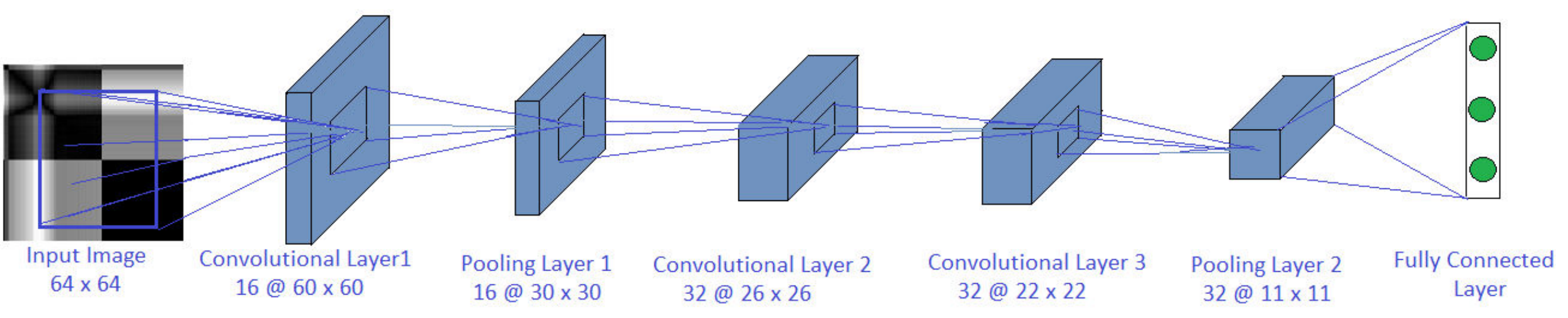}
	\caption{Architecture of CNN for Signal Image of size 64 x 64.}
	\label{fig:CNN Architecture}
\end{figure*}

\section{Materials and Methods}\label{sec:proposed method}

This section explains the proposed fusion frameworks called Multimodal Image Fusion (MIF) and Multimodal Feature Fusion (MFF). The common element in both of the proposed fusion framework is ECG signal to image transformation as shown in Figures~\ref{fig: MIF framework} and~\ref{fig: MFF framework}. Therefore in this section, first we will explain ECG signal to image transformation and then MIF, MFF and the two important elements of MFF, gated fusion network shown in Fig.~\ref{fig: gated fusion network} and architecture of CNN shown in Fig.~\ref{fig:CNN Architecture}, will be explained.

\subsection{ECG Signal to Image Transformation} \label{ sec:ECG to image formation}

For each fusion framework, we transform the input heart-beats into three types of images called GAF, RP and MTF images.

\subsubsection{Formation of Images by Gramian Angular Field (GAF)}\label{sec:gaf image formation}

Converting heart-beats of ECG into Gramian Angular Field (GAF) images maps the ECG in an angular coordinate system instead of typical rectangular coordinate system.

Consider that $E$ is an ECG signal of $n$ samples such that $E = \{s_1, s_2, s_3,...,s_k,s_l,...,s_n\}$. We normalized $E$ between 0 and 1 to get $\overline{E}$.
Now we map the normalized ECG in angular coordinate system by transforming the value into the angular cosine and the time stamps into the radius. Following equation is used to explain this encoding.

\begin{empheq}[right=\empheqrbrace]{equation}
\begin{split}
\beta = arccos(s_{k0}) \\
R = \frac{t_k}{C}
\end{split}
\end{empheq}
In the above equation, $s_{k0}$ is normalized $kth$ sample of the ECG, $t_k$ is the time stamp for $s_{k0}$ and $C$ is a constant to adjust the spread of the angular coordinate system.
This encoding provides two benefits. It is bijective and it conserves the spatial domain affiliations through the $R$~\cite{wang2015imaging}. Since the image location with respect to the ECG heart beat samples is consistent
along the principal diagonal, therefore, the original heart beat samples of ECG can be restored from angular coordinates~\cite{yang2020sensor}.

The angular viewpoint of the encoded image can be exploited by taking into account the sum/difference between each sample to indicate the correlation among various time stamps. The summation method, used in this article is explained by the following set of equations.

\begin{equation}
Grammian field  = cos(\beta_k + \beta_l)
\end{equation}

\begin{equation}\label{eq : GAF} 
Grammian field = \overline{E}^T . \overline{E} - {\sqrt{I - \overline{E}^2}}^T. \sqrt{I - {\overline{E}}^2}
\end{equation}
$I$ is the unit row vector in equation~\ref{eq : GAF}

\begin{figure*}[h]
	\centering
	\includegraphics[width=0.9\linewidth]{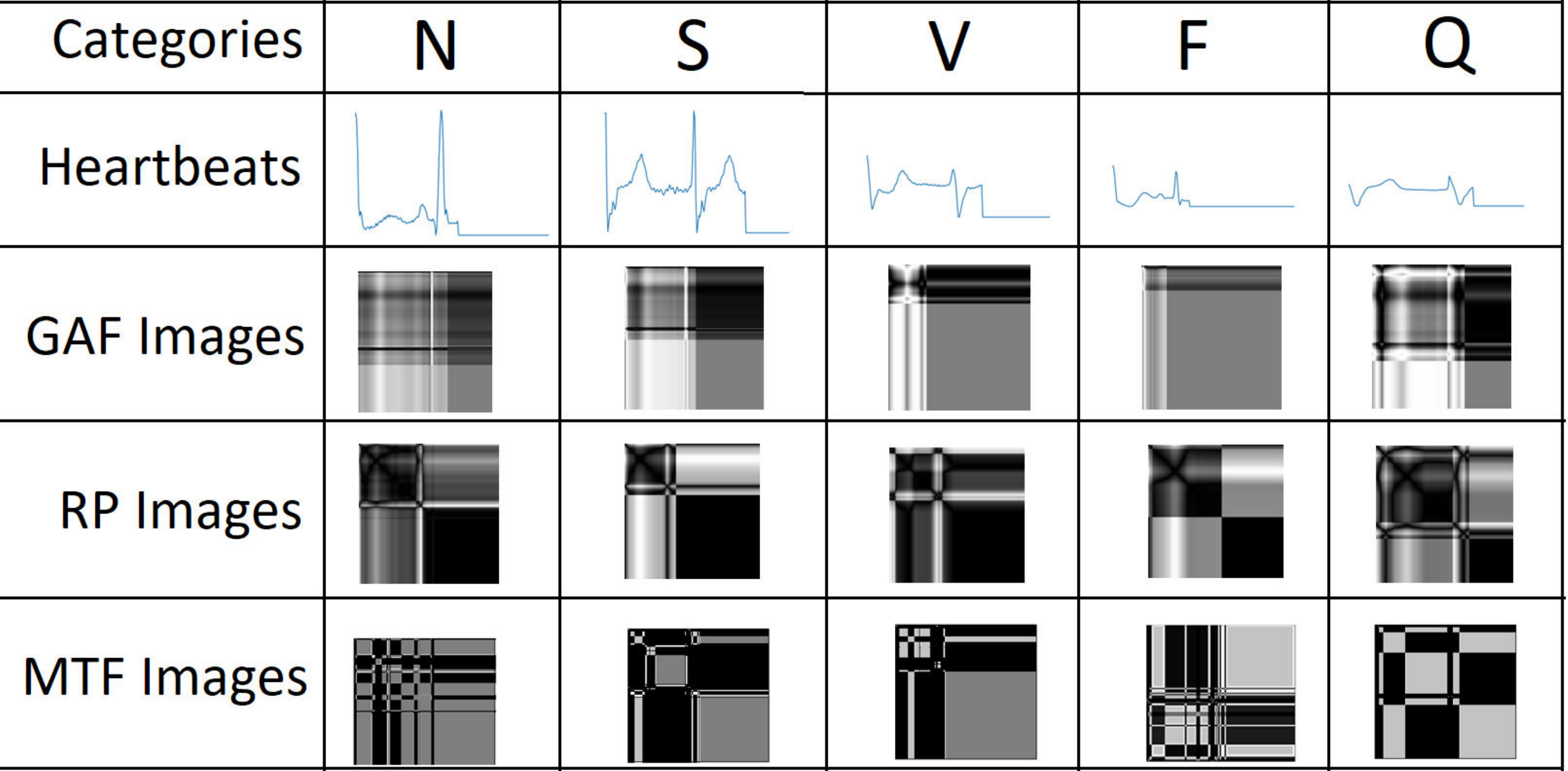}
	\caption{GAF, RP and MTF Images of MIT-BIH dataset according to the five different heartbeats defined in Table~\ref{tab : Mapping}}
	\label{fig: GAF Image}
\end{figure*}

GAF Images of five different categaories for MIT-BIH dataset are shown in Fig~\ref{fig: GAF Image}.

\subsubsection{Formation of Images by Recurrence Plot (RP)}\label{sec:rp image formation}

ECG is a non-stationary signal, therfore to visulaize the recurrent behavior and to observe the recurrence pattern of ECG signal~\cite{eckmann1995recurrence}, we encode ECG heartbeats into RP images. An RP image obtained from a heartbeat of ECG represents spacing between time points~\cite{blog}.

For ECG signal $E$ defined in section~\ref{sec:gaf image formation}, the recurrence plot is given by

\begin{equation}\label{ eq:RP}
R-plot = \alpha(\lambda - ||s(k) - s(l)||)
\end{equation}
where $\lambda$ is threshold and $\alpha$ is the heaviside function.

RP Images of five different categaories for MIT-BIH dataset are shown in Fig~\ref{fig: GAF Image}.

%

\subsubsection{ECG to Markov Transition Field (MTF) image conversion} \label{sec: mtf image formation}

For ECG heartbeats to MTF image encoding, we used the same approach explained in~\cite{wang2015encoding}. Let $E$ is the ECG signal defined in section~\ref{sec:gaf image formation}, then the foremost step is to define its $B$ bins based on quantiles and
assign every $s_k$ to the related bins $b_j (j\epsilon[1, B])$.
Second step is the construction of $B \times B$  weighted adjacency matrix $W$
by computing tranformations within quantile bins like
a first-order Markov chain on the time axis. Weighted adjacency matrix in the normalized form is called Markov transition matrix and is non-reative to the spatial domain characteristics, resulting in information loss. For handling the loss of information, Markov transition matrix is transformed to Markov transition field matrix (MTF) by stretching the transition likelihoods corresponding to the spatial domain locations. The MTF matrix is denoted by M and is shown below 

\begin{equation}\label{eq:Markov transition field matrix}
M=
\begin{bmatrix}
w_{lk|s_1\epsilon b_l,s_1\epsilon b_k}  & \dots & w_{lk|s_1\epsilon b_l,s_n\epsilon b_k}\\
w_{lk|s_2\epsilon b_l,s_1\epsilon b_k} & \dots  & w_{lk|s_2\epsilon b_l,s_n\epsilon b_k}\\
\vdots      &     \ddots &          \vdots    \\
w_{lk|s_n\epsilon b_l,s_1\epsilon b_k}  & \dots & w_{lk|s_n\epsilon b_l,s_n\epsilon b_k}

\end{bmatrix}
\end{equation}
Where $w_{lk}$ is the frequency of transition of a point between two quantiles. Since the formation of transformed matrix depends upon the chances of moving element, the MTF cannot be restored to original ECG signal.

Bins are the quantiles where the probability distribution is same. Any number of bins can be selected for ECG to MTF images. We decided to take 10 bins as the data is normalized between 0 and 1. These bins are defined during the formation of Weighted adjacency matrix which is the first step for creating MTF matrix shown in equation~\ref{eq:Markov transition field matrix}.

MTF Images of five different categaories for MIT-BIH dataset are shown in Fig~\ref{fig: GAF Image}.
%

For ECG to image transformation using GAT, RP and MTF methods, we are using the full length of heartbeats to transform 1D information to 2D. Therefore, ECG signal of any length can be transformed to images and then can be resized using interpolation.

We can see from Fig.~\ref{fig: GAF Image}, that for each kind of image (GAF, RP and MTF), the gray scale images are more interpretable. These images show different patterns for each of the five categories of MIT-BIH dataset. The x-y values of the 2D images are just pixel values of the GAF, RP, and MTF images.

\subsection{ Multimodal Image Fusion Framework}

Multimodal Image Fusion (MIF) framework is shown in Fig.~\ref{fig: MIF framework}. At the input, we transform the heartbeats of raw ECG signal into three types of images as described in section~\ref{ sec:ECG to image formation} and shown in Fig.~\ref{fig: GAF Image}. The motivation of choosing GAF, MTF and RP is that they are three different statistical methods of transforming ECG to images. During transformation they preserve the temporal information and hence they are lossless transformations. We combine these three gray scale images to form a triple channel image (GAF-RP-MTF).
A triple channel image is a colored image in which GAF, RP and MTF images are considered as three orthogonal channels like three different colors in RGB image space. However, this three-channel image is not conventional way of converting a gray scale image to RGB, rather in this paper all three gray scale images are formed from raw ECG data with different statistical methods. Thus, a three-channel image in the presented work carries statistical dynamics of the ECG and therefore, is more informative. Furthermore, three-channel image can be easily utilized with with off-the-shelf CNNs like AlexNet.

We use AlexNet, (CNN based model)~\cite{krizhevsky2012imagenet} for feature extraction and classification tasks and thus employ end-to-end deep learning where feature extraction and classification parts are embedded in a single network as shown in Fig.~\ref{fig: MIF framework}.

\subsection{Multimodal Feature Fusion Framework} \label{sec: multimodal feature fusion}

At the input of MFF, we transform ECG heartbeats into images as shown in Fig.~\ref{fig: MFF framework}. AlexNets are employed to learn features from input imaging modality. We extract these learned features from (fc-7) of each AlexNet and are then fused by an efficient Gated Fusion Network (GFN), backbone of the proposed MFF, which fuses the features effectively by taking care of their dimensionalities as well. These fused features are input of the SVM classifier as shown in Fig.~\ref{fig: MFF framework}.

\subsubsection{Gated Fusion Network} \label{sec: gated fusion network}

The architecture of our proposed gated fusion network (GFN) is shown in Fig.~\ref{fig: gated fusion network}. We have adapted this network from our previous work in~\cite{ahmad2020cnn}. The input to the GFN are the features extracted from the second last fully connected layer (fc-7) of each AlexNet as shown in Fig.~\ref{fig: MFF framework}.

Let $f_1$, $f_2$ and $f_3$ be the features from each imaging modality respectively. These feature are then convolved with high boost kernel $K$ as shown in Fig.~\ref{fig: gated fusion network}.

We used high boost filter for convolution with features since this filter precisely recognize important information of feature and accredits boosted value to every element of features according to its importance~\cite{mitchell2010image}. High boost filter is the difference between scaled version and low-pass version of the input image as shown below in equation~\ref{eq:highboostequation}.

\begin{equation}\label{eq:highboostequation}
f_{hb}(m,n) = cf(m,n) - f_{lp}(m,n)
\end{equation}

where $cf(m,n)$ and $f_{lp}(m,n)$ are respectively the scaled version and low pass version of image $f(m,n)$

In general, high boost filter is given by

\vspace{0.1cm}
\begin{equation}
K=
\begin{bmatrix}\label{eq:highboostmatrix}

-1 & -1 & -1 \\
-1 & c+8 & -1 \\
-1 & -1 & -1

\end{bmatrix}
\end{equation}	

where $c$ is the amplification factor that assigns the weights to the feature during convolution.

The best filter performance is obtained for $c$ = 1. Other values of $c$ produces less amplification. 

Thus, following high boost kernel is selected empirically that highlights the important characteristics. 

\vspace{0.1cm}
\begin{equation}\label{eq:final highboost eqn}
K=
\begin{bmatrix}

-1 & -1 & -1 \\
-1 & 9 & -1 \\
-1 & -1 & -1

\end{bmatrix}
\end{equation}	

High boost filter highlights the high frequency components while conserving the low frequency components. 

After convolution of features with the high boost filter, sigmoid function is used for generating proper gated weights $w_1$, $w_2$ and $w_3$ respectively as shown in Fig.~\ref{fig: gated fusion network}. 
Finally, we obtained point-wise product of the weights $w_1$, $w_2$ and $w_3$ and the features $f_1$, $f_2$ and $f_3$ respectively, to perform feature fusion and to generate fused features. The working of GFN can be understood by the following equations. 

\vspace{0.1cm}    

\begin{equation} \label{eq:first equation}
w_1 = \sigma(f_1 \circledast\ K)
\end{equation}

\begin{equation} \label{eq:second equation}
w_2 = \sigma(f_2 \circledast\ K)
\end{equation}

\begin{equation} \label{eq:third equation}
w_3 = \sigma(f_3 \circledast\ K)
\end{equation}

\begin{equation} \label{eq:fourth equation}
F_f(j)  = w_1 \odot f_1(j) + w_2 \odot f_2(j) + w_3 \odot f_3(j)
\end{equation}

\vspace{0.2cm}       

Where,
\vspace{0.1cm}

$\sigma(x) \triangleq \frac{1}{1+e^{-x}}$ : Sigmoid Function.

\vspace{0.1cm}

$a \circledast b$ : Convolution 
\vspace{0.1cm}

$a \odot b$ : Point Wise Multiplication

\vspace{0.1cm}

$F_i(k)$ : $k$th feature  of $i$th modality 

\vspace{0.1cm}

$F_f(k)$ : $kth$ Fused feature

%
%
%

\subsubsection{CNN Architecture}

Architecture of CNN used in proposed MFF is shown in Fig.~\ref{fig:CNN Architecture}. It consists of three convolutional layers, two pooling layers, and a fully connected layer. The first convolutional layer has 16 kernels of size 5x5, followed by pooling layer of size 2x2 and stride 2. Second and third convolutional layers have 32 kernels of size 5x5 followed by 2x2 pooling layer with stride 2.

\subsection{Classification Task and Classifier}

The classification task of the proposed methods is ECG heart beat classification for arrythmia and MI detection. 

The classification metrics used for classification are accuracy, precision and recall as shown in Tables~\ref{tab : experiments on MIT-BIH Dataset using AlexNet},~\ref{tab : experiments on MIT-BIH Dataset using CNN},~\ref{tab : experiments on PTB Dataset using AlexNet} and~\ref{tab : experiment results on PTB Dataset using CNN.}. The accuracies, precisions and recalls are calculated using following equations.

\begin{equation}
Accuracy = \frac{TP + TN}{TP + TN + FP + FN}
\end{equation}

\begin{equation}
Precision = \frac{TP}{TP + FP}
\end{equation}

\begin{equation}
Recall = \frac{TP}{TP + FN}
\end{equation}

where,

$TP$ = True positive 

$TN$ = True negative

$FP$ = False positive

$FN$ = False negative

We used Softmax classifier in proposed MIF and Support Vector Machines (SVM) classifier in proposed MFF for classification task.

Softmax classifier is a multiclass classifier or regressor used in the fields of machine learning.
Score function for softmax classifier computes the class specific probabilities whose sum is 1.

The mathematical representation of score function for softmax classifer is shown below.
\begin{equation}\label{eq:soft}
f(y)=\frac{e^{y_j}}{\sum_ke^{y_k}}
\end{equation}
where $y$ is the input vector and the score function maps the exponent domain to the probabilities.
 
In simplest form, the score function for SVM is the mapping of the input vector to the scores and is a simple matrix operation as shown in Equation~\ref{eq:svm}.
\begin{equation}\label{eq:svm}
f=Wx + b
\end{equation}
Where $x$ is the input vector, $W$ is the weight determined by input vector and the number of classes and $b$ is the bias vector.

\begin{table}
	\vspace{0.3cm}
	\centering
	
	\caption{Training Parameters for AlexNet and CNN}
	\label{tab:parameters for alexnet and CNN}
	\begin{tabular}{c c}
		
		\hline
		Training Parameters & Values   \\\hline 
		Momentum  &      0.9 \\
		Initial Learn Rate  &      0.005 \\
		Learn Rate Drop Factor  &      0.5 \\
		Learn Rate Drop Period  &      10 \\
		$L_2$ Regularization  &      0.004 \\
		MiniBatchSize  &   128 \\\hline		
		
	\end{tabular}
	
\end{table}

\subsection{Training and Optimization}

We resize images to 227 x 227 to perform experiments with AlexNet. We also perform experiments with smaller but computationally efficient CNN, whose architecture is shown in Fig.~\ref{fig:CNN Architecture}, to show that proposed frameworks can achieve comparable performance even with the smaller CNN. The comparison in terms of computational cost between both CNN models is provided in Table~\ref{tab:computationalcomparison}. We fine tune Alexnet by reducing the size of second last fully connected layer 'fc7' from 4096 to 512 and the size
of last fully connected layer ’fc8’ from 1000 to size equal to the
number of classes in our datasets. The size of “fc7” layer of AlexNet is 4096 which is according to size of classification layer which is 1000. For our MIT-BIH dataset and PTB dataset, we need the size of classification layer equal to 5 and 2 respectively due to number of classes in these datasets. Thus to make ‘fc7’ compatible with classification layer, we reduce its size to 512. The training parameters for AlexNet and CCN are shown in Table~\ref{tab:parameters for alexnet and CNN}.

For optimization of the deep networks, we used Stochastic Gradient Descent with Momentum (SGDM) algorithm. SGDM is a method which helps accelerate
gradients vectors in the right directions, thus leading to faster
converging. It is one of the most popular optimization algorithms
and many state-of-the-art models are trained using it.

\section{Experimental Results} \label{sec:Experiment and results}
\subsection {ECG Databases}

Experiments are performed with PhysioNet MIT-BIH Arrhythmia dataset~\cite{goldberger2000physiobank}~\cite{moody2001impact} for heartbeat classification and PTB Diagnostic ECG dataset~\cite{bousseljot1995nutzung} for MI classification using both proposed fusion frameworks. For experiments, ECG lead-II re-sampled data at sampling frequency of 125Hz is used as the input.

 We used the standardized form of both datasets provided in~\cite{blog2}. These datasets are already denoised and the training and testing parts are provided in the form of standard ECG heartbeats. Furthermore, five classes of arrythmia and MI localization has already been done and provided in terms of standard ECG heart-beats. Our study focused on ECG to image transformation and to the design of proposed multimodal fusion frameworks. The main focus is increasing the overall performance of classification of heartbeats. We did not attempt at modeling or solving for a specific type of noise.

We conduct our experiments on Matlab R2020a on a desktop computer with NVIDIA GTX-1070 GPU. 

 The experimental results are discussed in detail in section~\ref{sec:discussion on results}.
 
 \begin{table}[h]
 	
 	\centering 
 	
 	\caption{Mapping between annotations and AAMI EC57~\cite{association1998testing} categories}
 	\label{tab : Mapping}
 	\begin{adjustbox}{width=0.8\columnwidth,center}
 		\begin{tabular}{c c}
 			\hline
 			Categories    &Annotations   \\\hline
 			
 			\multirow{4}*{N}
 			& {Normal}       \\
 			& {Left/Right bundle branch block}   \\
 			& {Atrial escape}   \\
 			& {Nodal escape}   \\\hline
 			
 			\multirow{4}*{S}
 			& {Atrial Premature}       \\
 			& {Aberrant atrial premature}   \\
 			& {Nodal premature}   \\
 			& { Supra-ventricular premature}   \\\hline
 			
 			\multirow{2}*{V}
 			& {Premature ventricular contraction}       \\
 			& { Ventricular escape}   \\\hline
 			
 			\multirow{1}*{F}
 			& {Fusion of ventricular and normal}       \\\hline
 			
 			\multirow{3}*{Q}
 			& {Paced}       \\
 			& {Fusion of paced and normal}   \\
 			& {Unclassifiable}   \\\hline
 		\end{tabular}
 	\end{adjustbox}
 	
 \end{table}

 \begin{table}[h]
 	
 	\caption{Information about Number of Heartbeats before and after SMOTE for training component of MIT-BIH Dataset}
 	\label{tab : dataset heartbeat information}
 	
 	\begin{adjustbox}{width=\columnwidth,center}
 		\begin{tabular}{c c c c}
 			\hline
 			Dataset     & \makecell{Classes}    &\makecell{Original \\ heartbeats}    & \makecell{Number of heartbeats \\ after SMOTE}     \\\hline

 			\multirow{5}*{MIT-BIH}
 			
 			& N    & 72471    & 72471    \\
 			& S    & 2223   & 30000    \\
 			& V    & 5788    & 20000    \\
 			& F    & 641    & 20000    \\
 			& Q    & 6431    & 10000    \\\hline

 		\end{tabular}
 	\end{adjustbox}
 	
 \end{table}

 \begin{table}
 	
 	\caption{Training and Testing Samples of datasets}
 	\label{tab : Training and testing samples}
 	
 	\centering 
 	\begin{adjustbox}{width=\columnwidth,center}
 		\begin{tabular}{ c c c }
 			\hline
 			Dataset    &Training Samples    & Testing Samples     \\\hline
 			MIT-BIH    & 152471   & 21892    \\
 			PTB    & 11641   & 2911    \\\hline

 		\end{tabular}
 	\end{adjustbox}
 	
 \end{table}

\subsubsection{PhysioNet MIT-BIH Arrhythmia Dataset} \label{sec : physionet data}

Forty seven subjects were involved during the collection of ECG signals for the dataset. The data was collected at the sampling rate of 360Hz and each beat is annotated by at least two experts. Using these annotations, five different beat categories are created in accordance with Association for the Advancement
of Medical Instrumentation (AAMI) EC57 standard~\cite{association1998testing} as shown in Table~\ref{tab : Mapping}.

For training on CNN, we need large number of samples. We use the same testing and training segments provided in~\cite{blog2} to train on CNNs. Since there is a class-imbalanced in the training part of the dataset as apparent from the numbers, we applied SMOTE~\cite{chawla2002smote} to upsample the minority classes (classes other than N) and finally settled on the numbers shown in the right column of Table~\ref{tab : dataset heartbeat information}.

SMOTE is a data augmented technique which is used to reduce overfitting during training and is helpful to reduce the biasness of classifier.

We perform experiments using both proposed fusion frameworks on MIT-BIH dataset with the
training and testing samples shown in Table~\ref{tab : Training and testing samples} and with the training parameters shown in Tables~\ref{tab:parameters for alexnet and CNN}. The experimental results are shown in Tables~\ref{tab : experiments on MIT-BIH Dataset using AlexNet} and~\ref{tab : experiments on MIT-BIH Dataset using CNN}.

\begin{table}
	\centering
	
	\caption{Experimental results of MIT-BIH Dataset using AlexNet.}
	\label{tab : experiments on MIT-BIH Dataset using AlexNet}
	\scalebox{1}{
		\begin{tabular}{c c c c }	
			\hline
			Modalities & Accuracies\% & Precision\% & Recall\%  \\\hline
			GAF Images only  &  97.3 & 85 & 91 \\
			RP Images only & 97.2 &  82 &  93  \\
			MTF Images only & 91.5 & 86 &  89  \\	
			Concatenation Fusion & 97 &  82 & 91 \\
			Average Fusion & 98.5 &  95 & 93.1 \\	
			Proposed MIF  &  98.6 & 93 & 92 \\
			Proposed MFF  & 99.7 & 98 & 98 \\\hline
			
	\end{tabular}}
	
\end{table} 

\begin{table}
	
	\caption{Experimental results of MIT-BIH Dataset using simpler CNN of Fig.~\ref{fig:CNN Architecture}}.
	\label{tab : experiments on MIT-BIH Dataset using CNN}
	\centering
	\scalebox{1}{
		\begin{tabular}{c c c c }	
			\hline
			Modalities & Accuracies\% & Precision\% & Recall\%  \\\hline
			GAF Images(gray scale)  &  94.2  & 74.2  &  91   \\
			RP Images(gray scale) & 96.3  & 80  & 90   \\
			MTF Images(gray scale) & 94  & 72  &  86  \\	
			Concatenation Fusion & 94.6  &  80.4  &  84  \\
			Average Fusion & 97.6  &  87  & 92   \\	
			Proposed MFF  & 98.3  &  90.5  &  93  \\\hline
			
	\end{tabular}}

\end{table} 

\begin{table}
	\centering
	
	\caption{Experimental results of PTB Dataset using AlexNet.}
	\label{tab : experiments on PTB Dataset using AlexNet}
	\scalebox{1}{
		\begin{tabular}{c c c c }	
			\hline
			Modalities & Accuracies\% & Precision\% & Recall\%  \\\hline
			GAF Images only  &  98.4 & 98 & 96  \\
			RP Images only & 98 &  98 & 94  \\
			MTF Images only & 95.3 & 94 & 89 \\	
			Concatenation Fusion & 97.4 & 95 & 95  \\
			Average Fusion & 98.5& 97 & 98  \\
			Proposed MIF &  98.4 & 98 & 94  \\	
			Proposed MFF  & 99.2 & 98 & 98 \\\hline
			
	\end{tabular}}
	
\end{table}

\begin{table}
	\centering
	
	\caption{Experimental results of PTB Dataset using simpler CNN of Fig.~\ref{fig:CNN Architecture}}.
	\label{tab : experiment results on PTB Dataset using CNN.}
	\scalebox{1}{
		\begin{tabular}{c c c c}	
			\hline
			Modalities & Accuracies\% & Precision\% & Recall\% \\\hline
			GAF Images (gray scale)  &  94.7 & 91 &  90  \\
			RP Images (gray scale) & 95.1 & 95 & 87  \\
			MTF Images(gray scale) & 86.6 &  80 & 69 \\	
			Concatenation Fusion & 92.2 & 88 & 84 \\
			Average Fusion & 96.3 & 91 & 94 \\	
			Proposed MFF  & 96.5 & 94 & 93  \\\hline
			
	\end{tabular}}
	
\end{table}

\subsubsection{PTB Diagnostic ECG dataset}

Two hundred and ninety (290) subjects took part during collection of ECG records for PTB Diagnostics dataset. 148 of them are diagnosed as MI, 52 healthy control, and the rest are diagnosed with 7 different diseases. Frequency of 100Hz is used for each ECG record from 12 leads. However, for our experiments, we used lead II ECG recordings and worked with healthy control and MI categories. 

We perform experiments using both proposed fusion frameworks on PTB dataset with training and testing samples shown in Table~\ref{tab : Training and testing samples} and with training parameters shown in Tables~\ref{tab:parameters for alexnet and CNN}. Training and testing parts of the dataset are provided in~\cite{blog2} to train CNN models. The experimental results are shown in Tables~\ref{tab : experiments on PTB Dataset using AlexNet} and~\ref{tab : experiment results on PTB Dataset using CNN.}

\section{Discussion}\label{sec:discussion on results}

We present the comparative results of the proposed frameworks with the state-of-the art methods in  Tables~\ref{tab : Comparison of MIT-BIH} and~\ref{tab : Comparison of PTB}. As we can see, our proposed frameworks considerably outperform the existing methods in terms of accuracy, precision, and recall.

To justify the importance of the proposed fusion frameworks, we assess the performance of different components of the proposed framework with both datasets by concatenation and average fusion methods. We performed average fusion  by accrediting the unity value to all the weights i.e $w_1$ = 1, $w_2$ = 1  and $w_3$ = 1 in the gated fusion network. Since we have three modalities, therefore, by taking simple average, we get the equal value of 0.333 for each weight. We also experiment with 0.333 and get the same results. Since weights are equal in average fusion, therefore, to make things simpler, we assign a unity value to every weight. It is possible that better weight can be acquired through trainable weight coefficients. This is something we plan to investigate in future. Tables~\ref{tab : experiments on MIT-BIH Dataset using AlexNet},~\ref{tab : experiments on MIT-BIH Dataset using CNN},~\ref{tab : experiments on PTB Dataset using AlexNet} and~\ref{tab : experiment results on PTB Dataset using CNN.} reports the results of assessing different fusion methods along with proposed fusion frameworks.

\begin{table}
	\centering
	\caption{Comparison of heart beat Classification results of MITBIH Dataset with Previous Methods}
	\label{tab : Comparison of MIT-BIH}
	\scalebox{1}{
		\begin{tabular}{c c c c}	
			\hline
			Previous Methods & Accuracies\% & Precision\% & Recall\% \\\hline
			
			Izci et al.~\cite{izci2019cardiac} & 97.96 & - & -  \\
			Dang et al.~\cite{dang2019novel} & 95.48 & 96.53 & 87.74  \\
			Li et al.~\cite{li2019automated} & 99.5 & 97.3 & 98.1 \\
			Zhao et al.~\cite{zhao2017electrocardiograph} & 98.25 & - & - \\
			Oliveria et al.~\cite{oliveira2019novel} & 95.3 & - & - \\	
			Huang et al.~\cite{huang2019ecg} & 99 & - & - \\
			Shaker et al.~\cite{shaker2020generalization} & 98 & 90 & 97.7 \\
			Kachuee et al.~\cite{kachuee2018ecg} & 93.4 & - & - \\
			Xu et al.~\cite{xu2020interpretation} & 95.9 & - & - \\
			He et al.~\cite{he2020automatic} & 98.3 & - & - \\
			Qiao et al.~\cite{qiao2020fast} & 99.3 & - & - \\
			\textbf{Proposed MIF} & \textbf{98.6} & \textbf{93} & \textbf{92} \\	
			\textbf{Proposed MFF} & \textbf{99.7} & \textbf{98} & \textbf{98}  \\\hline		
	\end{tabular}}
	
\end{table} 

\begin{table}
	\centering
	\caption{Comparison of MI Classification results of PTB Dataset with Previous Methods}
	\label{tab : Comparison of PTB}
	\begin{tabular}{c c c c}	
		\hline
		Previous Methods & Accuracies\% & Precision\% & Recall\% \\\hline
		
		Dicker et al.~\cite{diker2019novel} & 83.82 & 82 & 95  \\
		Acharya et al.~\cite{acharya2017application} & 95.22 & 95.49 & 94.19  \\
		Kojuri et al.~\cite{kojuri2015prediction} & 95.6 & 97.9 & 93.3  \\
		Kachuee et al.~\cite{kachuee2018ecg} & 95.9 & 95.2 & 95.1 \\
		Liu et al.~\cite{liu2017real} & 96 & 97.37 & 95.4 \\
		Sharma et al.~\cite{sharma2015multiscale} & 96 & 99 & 93 \\
		Chen et al.~\cite{chen2018multi} & 96.18 &  97.32 & 93.67 \\
		Cao et al.~\cite{cao2020multi} & 96.65 &  -  & - \\
		Ahamed et al.~\cite{ahamed2020ecg} & 97.66 &  -  & - \\
		\textbf{Proposed MIF} & \textbf{98.4} & \textbf{98} & \textbf{94} \\	
		\textbf{Proposed MFF} & \textbf{99.2} & \textbf{98} & \textbf{98}  \\\hline		
	\end{tabular}
	
\end{table}

The performance of concatenation fusion is poor as compared to other methods as shown by experimental results. Concatenation fusion creates high dimensional feature vector that leads to the additional computational cost and deterioration of information during classification~\cite{akbas2007automatic}.

We also provide the comparison of both proposed fusion frameworks in terms of inference speed as shown in Table~\ref{tab : Inference speed}. Inference speed is the time consumed by classifier to recognize one test sample. It is expressed in microseconds ($\mu$s). It is observed that MFF yields high accuracy, precision and recall for both datasets as compared to MIF, however, MIF is computationally efficient in terms of inference speed.

\begin{table}
	\centering
	
	\caption{Comparison of Computational Cost of AlexNet and CNN of Fig.~\ref{fig:CNN Architecture} using MFF Framework on MIT-BIH Dataset}
	\label{tab:computationalcomparison}
	
	\begin{adjustbox}{width=0.9\columnwidth,center}
		\begin{tabular}{c c c}
			
			\hline
			CNN Model & Fusion Framework & Training Parameters  \\\hline
			AlexNet & MFF  & 9259427  \\	
			AlexNet & MIF  & 3086475  \\
			CNN of Fig.~\ref{fig:CNN Architecture} & MFF & 612069  \\\hline
		\end{tabular}
	\end{adjustbox}
	
\end{table}

Since we experiment with two different CNNs, we provide comparison between both CNNs in terms of computational cost as shown in Table~\ref{tab:computationalcomparison}. Since there is a trade off between accuracy and computational cost, we observe from Tables~\ref{tab : experiments on MIT-BIH Dataset using AlexNet},~\ref{tab : experiments on MIT-BIH Dataset using CNN} and~\ref{tab:computationalcomparison} that CNN, shown in Fig.~\ref{fig:CNN Architecture}, is less accurate than AlexNet but is computationally efficient. 

\begin{table}
	
	\centering 
	
	\caption{Comparison of Inference Speed of both Proposed Fusion Frameworks using AlexNet.}
	\label{tab : Inference speed}
	\begin{adjustbox}{width=\columnwidth,center}
		\renewcommand\arraystretch{1.3}
		\begin{tabular}{ c c c}
			\hline
			Dataset   & Fusion Method   & \makecell{Inference Speed($\mu$s)}     \\\hline
			
			\multirow{2}*{MIT-BIH}
			& Multimodal Image Fusion    & 1233     \\
			
			& \multirow{1}*{Multimodal Feature Fusion}
			& 1670      \\	
			\hline
			\multirow{2}*{PTB}
			& Multimodal Image Fusion   & 1205     \\
			
			& \multirow{1}*{Multimodal Feature Fusion}
			& 1470       \\\hline	
			
		\end{tabular}
	\end{adjustbox}
	
\end{table}

We prefer SVM classifier over softmax classifier since we have experimentally proved in our previous work~\cite{ahmad2018towards} that SVM performs better than softmax, which is typically built into any CNN framework.
Softmax classifier reduces the cross entropy function while SVM employs a margin based function. The more rigorous nature of classification is the reason of better performance of SVM over softmax.

The comparison  provided in Tables~\ref{tab : Comparison of MIT-BIH} and~\ref{tab : Comparison of PTB} is on the basis of datasets and the performance metrics. There are slight changes in the conditions for testing in few of the comparisons, However, it is appropriate to compare the results.

The limitation of the proposed Multimodal Image Fusion (MIF) Framework is that it requires exactly three different statistical gray scale images for creating a triple channel compound image.
Since Multimodal Feature Fusion (MFF) Framework is using three separate AlexNet for training on GAF, RP and MTF images, it requires more time for training and inference.

\section{Conclusion}

We proposed two computationally efficient multimodal fusion frameworks for ECG heart beat classification called Multimodal Image Fusion (MIF) and Multimodal Feature Fusion (MFF). At the input of these frameworks, we convert ECG signal into three types of images using Gramian Angular Field (GAF), Recurrence Plot (RP) and Markov Transition Field (MTF). In MIF, we first perform image fusion by combining three input images to create a three channel single image which used as input to the CNN. In MFF, highly informative cues are pulled out from penultimate layer of CNN and they are fused and used as input for the SVM classifier. We demonstrate the superiority of the proposed fusion frameworks by performing experiments on PhysionNet’s MIT-BIH for five different arrhythmias and on PTB diagnostics dataset for MI classification. Experimental results prove that we beat the previous state-of-the-art in terms of classification accuracy, precision and recall. The important finding of this study is that the multimodal fusion of modalities increases the performance of the machine learning task as compare to use the modalities individually.

\bibliographystyle{IEEEtran}

\begin{IEEEbiography}[{\includegraphics[width=1in,height=1.25in,clip,keepaspectratio]{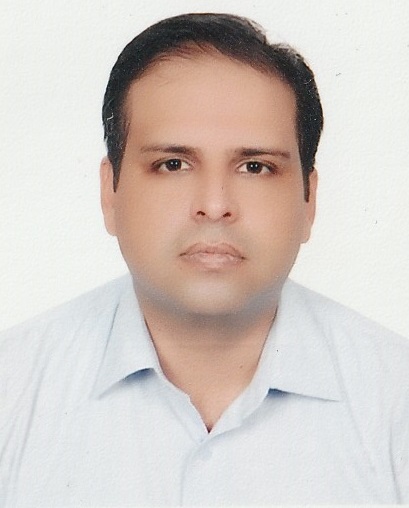}}]{Zeeshan Ahmad} received B.Eng. degree in Electrical Engineering from NED University of Engineering and Technology Karachi, Pakistan in 2001, M.Sc. degree in Electrical Engineering from National University of Sciences and Technology Pakistan in 2005 and MEng. degree in Electrical and Computer Engineering from Ryerson University, Toronto, Canada in 2017.  He is currently pursuing Ph.D. degree with the Department of Electrical and Computer Engineering, Ryerson University, Toronto, Canada. His research interests include Machine learning, Computer vision, Multimodal fusion, signal and image processing.
\end{IEEEbiography}

\begin{IEEEbiography}[{\includegraphics[width=1in,height=1.25in,clip,keepaspectratio]{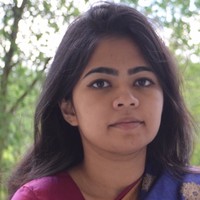}}]{Anika Tabassum} is a recent graduate of the MSc Data Science and Analytics program at Ryerson University. She received her BA degree in Computer Science from McGill University in 2013. She has previously worked 4+ years as a software engineer/developer.
\end{IEEEbiography}

\begin{IEEEbiography}[{\includegraphics[width=1in,height=1.25in,clip,keepaspectratio]{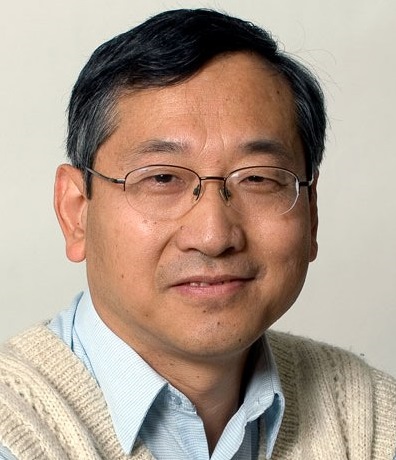}}]{Dr. Ling Guan} is a professor of Electrical and Computer Engineering at
	Ryerson University, Toronto, Canada, and was a Tier I Canada Research
	Chair in Multimedia and Computer Technology from 2001 to 2015. Dr. Guan has published extensively in multimedia processing and communications, human-centered computing,
	machine learning, adaptive image and signal processing, and, more
	recently, multimedia computing in immersive environment. He is a Fellow of
	the IEEE, an Elected Member of the Canadian Academy of Engineering, and an
	IEEE Circuits and System Society Distinguished Lecturer. 
\end{IEEEbiography}

\begin{IEEEbiography}[{\includegraphics[width=1in,height=1.25in,clip,keepaspectratio]{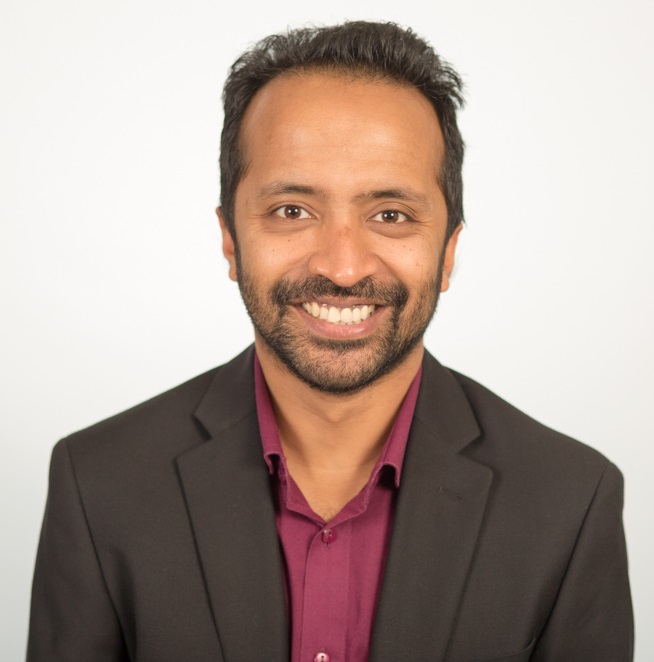}}]{Naimul Khan} is an assistant professor of Electrical and Computer Engineering at Ryerson University, where he co-directs the Ryerson Multimedia Research Laboratory (RML). His research focuses on creating user-centric intelligent systems through the combination of novel machine learning and human-computer interaction mechanisms.  He is a recipient of the best paper award at the IEEE International Symposium on Multimedia, the OCE TalentEdge Postdoctoral Fellowship, and the Ontario Graduate Scholarship. He is a senior member of IEEE and a member of ACM.
\end{IEEEbiography}

\EOD

\end{document}